\documentclass[final]{cvpr}

\usepackage{times}
\usepackage{epsfig}
\usepackage{graphicx}
\usepackage{amsmath}
\usepackage{amssymb}
\usepackage{multirow}
\usepackage{color}
\usepackage{colortbl}
\usepackage[colorlinks,linkcolor=blue,breaklinks=true,bookmarks=false]{hyperref}

\def\confYear{CVPR 2021}
\newcommand*{\affaddr}[1]{#1}

\begin{document}

\title{CLIP2Video: Mastering Video-Text Retrieval via Image CLIP}

\author{
    Han Fang\thanks {This work is done when Han Fang is an intern at Tencent} \quad 
    Pengfei Xiong\footnotemark[1] \thanks {Corresponding author} \quad 
    Luhui Xu \quad  
    Yu Chen \quad \\\\

\affaddr{PCG, Tencent} \\

\tt\small fanghan@bupt.edu.cn, xiongpengfei2019@gmail.com, \{lukenxu, andyyuchen\}@tencent.com \\
\tt\small \url{https://github.com/CryhanFang/CLIP2Video}
}
\maketitle

\begin{abstract}

We present \textit{CLIP2Video} network to transfer the image-language pre-training model to video-text retrieval in an end-to-end manner. Leading approaches in the domain of video-and-language learning try to distill the spatio-temporal video features and multi-modal interaction between videos and languages from a large-scale video-text dataset.
Different from them, we leverage pretrained image-language model, simplify it as a two-stage framework with co-learning of image-text and enhancing temporal relations between video frames and video-text respectively, make it able to train on comparatively small datasets. 
Specifically, based on the spatial semantics captured by Contrastive Language-Image Pretraining (CLIP) model, our model involves a Temporal Difference Block to capture motions at fine temporal video frames, and a Temporal Alignment Block to re-align the tokens of video clips and phrases  and enhance the multi-modal correlation.
We conduct thorough ablation studies, and achieve state-of-the-art performance on major text-to-video and video-to-text retrieval benchmarks, including new records of retrieval accuracy on MSR-VTT, MSVD and VATEX.

\end{abstract}


\section{Introduction}



Video-text retrieval, which aims to return for a given textual query the most relevant videos, is a fundamental research task for multi-modal video-and-language understanding. It becomes an emerging requirement with the increasing of web videos. In the past years, remarkable progress \cite{yu2018joint, zhu2020actbert, lei2021less, liu2019use, gabeur2020multi, dzabraev2021mdmmt, mithun2018learning, zhang2018cross, liu2021hit, dong2019dual, TimeSformer, arnab2021vivit} has occurred across many video-text benchmarks \cite{caba2015activitynet, xu2016msr, wang2019vatex, chen2011collecting, Rohrbach2015lsmdc, Hendricks2017DiDeMo}. 

Most such approaches focus on two critical issues. The first is the visual feature representation in the video domain. Different from image, video feature representation considers both spatial and temporal dimensions. Multi-path 2D or 3D convolutional networks \cite{zhang2018s3d,feichtenhofer2019slowfast,wang2019hallucinating} are still the core operators for feature learning, while both the spatial and temporal representations are considered in the same convolution operation for semantic and motion modalities. The other one is multi-modal interaction between video and languages. Based on a large-scale video-text dataset, single-stream or two-stream methods \cite{zhu2020actbert, lei2021less, bain2021frozen, dzabraev2021mdmmt, zhang2018cross, dong2019dual} are adopted to jointly train video features and text features inside the same embedding space. Nevertheless, these two problems are complex enough to make it difficult to achieve both goals in the same network. Some massive pre-training video-text datasets are sorted out to solve this problem, \eg Howto100M \cite{miech2019howto100m}. However, the pretrained models show limited performance gain for video-text retrieval, while annotated video data is hard to collect. 


To address these challenges, we rethink the video-text retrieval task from a more macroscopic point of view. While videos and sentences are both sequential, the meaning of a word can be reflected in an image or a sequence frames. For example, atomic actions need to be contextualized with short-term segments, while object is described in single image. Thus, the video-and-language understanding is divided into two independent problems, spatial representation of multi-modal image-text training, and temporal relationships of video frames and video-language. Compared with the video-text pre-training model, the learning of image-text model is much easier. The prominent success of the CLIP \cite{radford2021learning} (Contrastive Language-Image Pre-training) has demonstrated its capability of learning SOTA image representations from linguistic supervision with pre-training on large-scale image and text pairs.

\begin{figure*}
        \begin{center}
            \includegraphics[width=0.9\linewidth]{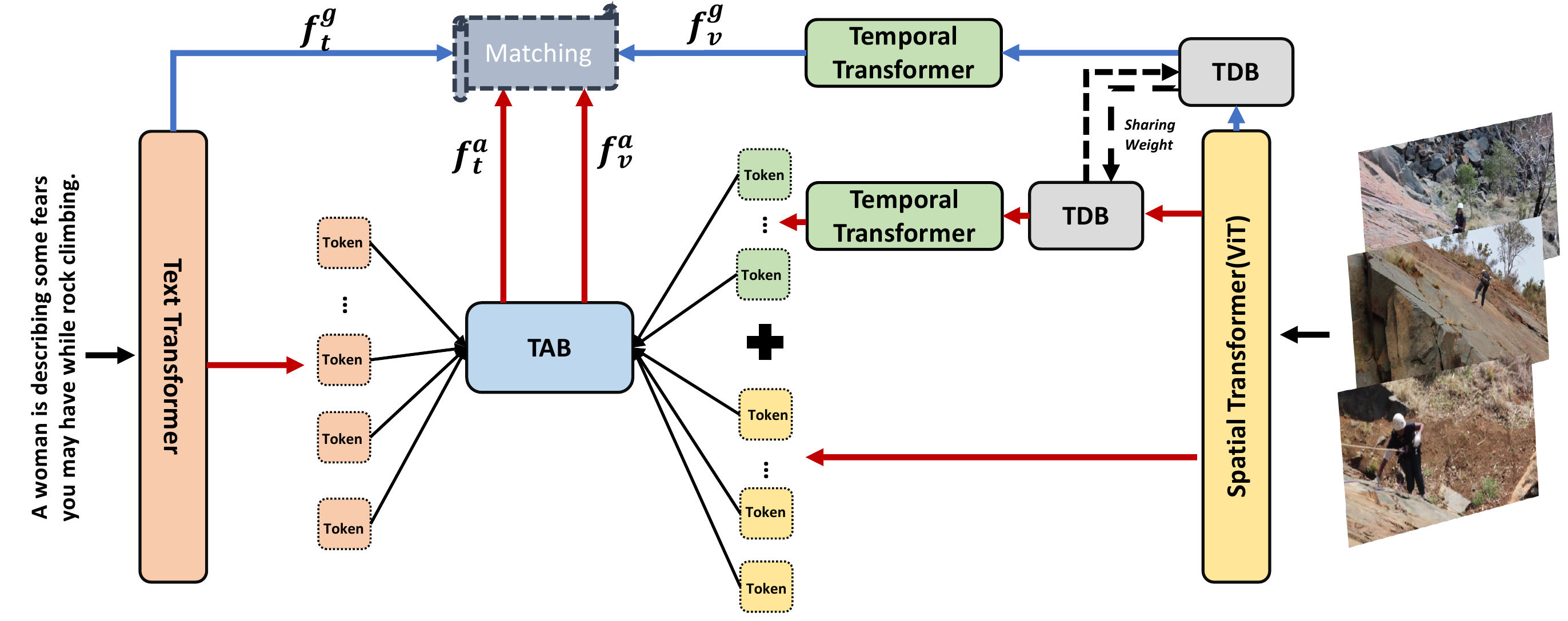}
        \end{center}
        \caption{Overview of CLIP2Video. It consists of two key components: Temporal Difference Block (TDB), which is used to enhance temporal interaction between frames; Temporal Alignment Block (TAB), which is adopted to align video clips and contextual words in the same space, capturing the motion change by cross-modal understanding.
        }
        \label{pipeline}        
\end{figure*}

Based on the spatial semantics captured by CLIP \cite{radford2021learning}, we present \textit{CLIP2Video} network to transfer the image-language pre-training model to video-text retrieval with two parts: Temporal Difference Block (TDB) and Temporal Alignment Block (TAB). As their names imply, the two components are devised to confront with the temporal relations of video frames and video-language respectively. We transform the video feature into a feature sequence. For Temporal Difference Block, we add the difference of image frames to the sequence to simulate the motion change. In respect of Temporal Alignment Block, the video sequence and text sequence are aligned to the same space to enhance the correlation between video clips and phrases. Fig.\ref{pipeline} shows the structure of these two components. Similar to our work, the concurrent works from \cite{portillo2021straightforward,luo2021clip4clip} are also built on CLIP for video-text retrieval. However, both of these two works only analyze a series of experiments to verify the effects of CLIP model for pre-training. Instead, we further study how to better model the temporal dependency between video frames and video-text, taking advantage of existing remarkable image pretrained model.

In summary, there are three main contributions in our paper: 
\begin{itemize}
    \item We put forward a new perspective of video-language learning with two independent modules, image-text multi-modal learning and temporal relationships between video frames and video-text, which respectively solves the multi-modal learning problems in the spatial and temporal aspects. 
    \item We introduce two modules, Temporal Difference Block and Temporal Alignment Block,  handling with temporal relationship of video frame and video text respectively, which can be used for any video language problem.
    \item We report new records of retrieval accuracy on several text-to-video and video-to-text retrieval benchmarks, including MSR-VTT \cite{xu2016msr}, MSVD \cite{chen2011collecting} and VATEX \cite{wang2019vatex}. Accompanied by thorough ablation studies, the large improvements are pin-pointed as contributions by our divided concept.
\end{itemize}

\section{Related Work}

\textbf{Video Representation Learning.} 
Previous works mainly focus on 2D/3D spatial-temporal convolution for video representation. SlowFast \cite{feichtenhofer2019slowfast} explores a network architecture that operates in two pathways and different frame rates. 
Recently ViT \cite{dosovitskiy2020image}, a transformer-based image encoder, which is shown to deliver impressive performance on image categorization, has attracted much attention. While introduced into video domain, ViViT \cite{arnab2021vivit,sharir2021image} and time transformer \cite{TimeSformer} propose several variants of ViT, including those that are more efficient by factorising the spatial and temporal dimensions of the input video. Similar to our work, they have been carried out this idea that adopt separately attentions on temporal and spatial with two-path transformer models. However, their methods still focus on designing an end-to-end network structure to decouple the two problems. We mainly investigate the effective temporal relationships based on multi-modal image-text learning.

\textbf{Video-language Learning.} Learning visual representation from text representation is an emerging research topic with the benefit of large-scale visual and language pairs collection. Howto100M \cite{miech2019howto100m} is one of the largest datasets for video-text multi-modal pretraining. However, there exists much ambiguity noises between text semantics and video content. MIL-NCE \cite{miech2020end} mainly investigates to leverage this noisy instructional videos to learn a better video encoder in an multi-modal learning manner. 
Others \cite{lei2021less, Deepti2019large, Jonathan2020Web, Tianhao2020cpd} collect videos and accompanied text information from  YouTube and Instagram to to learn spatio-temporal features in an efficient weakly-supervised manner. However, compared with image text pretraining data set, the collection of video text data set is much more complex, and its noise is also much larger. This makes the video pretraining model difficult to play a very big role. 

\textbf{Video-Text Retrieval.} 
Early works on video-text retrieval designed intensive fusion mechanisms
for cross-modal learning. Based on a large-scale annotated video-text dataset, single-stream or two-stream methods \cite{zhu2020actbert, lei2021less, bain2021frozen, dzabraev2021mdmmt, zhang2018cross, dong2019dual} are adopted to jointly extract video features and language features and project them into the same embedding space. Recently, the pre-trained models have dominated the leaderboard of the video-text retrieval with noticeable results on zero-shot retrieval. Concurrent to our work, \cite{portillo2021straightforward} apply CLIP for
zero-shot prediction. We propose to directly transfer the powerful knowledge from the pre-trained CLIP and continue learn the designed video-based CLIP2video model on a video-language dataset. Empirical studies present the effectiveness of the CLIP2video model.

\section{Methodology}

Given a set of captions as queries, our goal is to search for the corresponding videos by mapping video and text into joint embedding space. Inspired by the success of transferring image-text pre-training knowledge into video-text learning \cite{lei2021less}, we directly adopt CLIP \cite{radford2021learning} for initialization to extend the ability in text-to-video retrieval. Different from image-to-text retrieval, temporal correlations of visual clues fully reflect the semantics of video, which helps to facilitate cross-modal understanding. 
So, a temporal difference block is proposed to excite the motion-sensitive interactions explicitly. Meanwhile, we propose temporal alignment block to fully exploit the alignment between context of text and content of key frames.

\subsection{Temporal Difference Block}
To obtain video embedding, vision transformer (ViT) \cite{dosovitskiy2020image} is adopted firstly to encode every frame into feature. In particular, ViT extracts $N$ non-overlapping image patches and perform linear projection to map every patch into $1D$ token. With injection of positional embedding and extra [CLS] token, the sequence of tokens $\mathfrak{z}$ are input into $L_s$-layer transformer to model the correlation of each patch, where each layer $l_s$ comprises of Multi-Head Self-Attention ($MSA$) \cite{vaswani2017attention}, layer normalization ($LN$) \cite{ba2016layer}, and Multi-layer Perception ($MLP$). Then, a linear projection is adopted to encode $\mathfrak{z}_{cls}^{L_s}$ into embedding of the same dimension as text embedding for frame representation. 
\begin{figure}
        \begin{center}
            \includegraphics[width=1\linewidth]{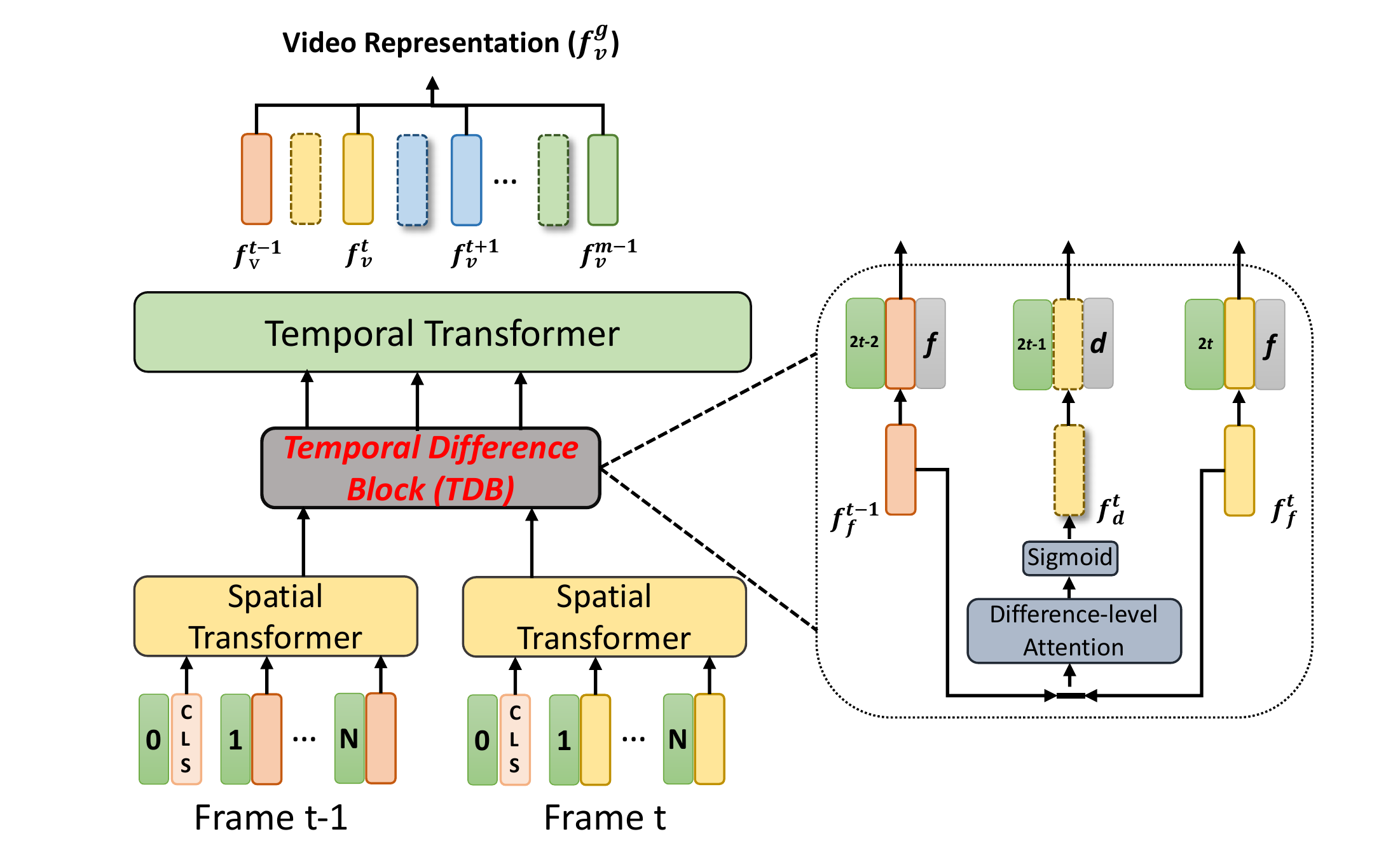}
        \end{center}
        \caption{Temporal Difference Block. By inserting TDB in adjacent frames, the motion can be explicitly provided for temporal transformer to capture temporal representation.}
        \label{TDB}  
\end{figure}

However, as shown in Fig.\ref{pipeline}, the spatial ViT models correlation within each frame without consideration of temporal indices. So, to exploit interactions between different frames, we propose a $L_t$-layer temporal transformer to encode video representation. Frame embeddings output by ViT are concatenated as frame tokens. Since two successive frames contain content displacement, which reflects the actual actions, we explicitly propose 
temporal difference block to extend the input and guide temporal transformer to encode more motion-related representations. The structure is shown in Fig.\ref{TDB}. Specifically, we adopt transformed difference of frame embedding between adjacent time stamps to describe the motion change, which is formulated as:
\begin{equation}
\begin{split}
\label{differentailEnhancedInput}
\vspace{-0.2cm}
\mathbf{F_d} &= 2\zeta(\delta(\{f_{f}^1 - f_{f}^0, f_{f}^2 - f_{f}^1, ..., f_{f}^{m-1} - f_{f}^{m-2}\} + \textbf{P})) - 1
\end{split}
\end{equation}

where $\textbf{P}$ is the positional embedding, $f_{f}^{m-1}$ and $f_{f}^{m-2}$ are the two adjacent frame embeddings, $\zeta$ is the sigmoid function, $\delta$ is 1-layer transformer, and $\mathbf{F_d}$ is the difference-enhanced token. Instead of  adopting  subtraction directly to represent difference, we propose to perform difference-level attention $\delta$  with sigmoid transformation. By employing attention transformation on the whole subtraction, the subtraction of successive frame embedding can be encoded to mode long-term relationship of all segments and normalized into [-1, 1] to indicate difference. Then we insert difference-enhanced tokens between every adjacent frames as:
\begin{equation}
\begin{split}
\label{MSA}
\vspace{-0.4cm}
\mathbf{F_{te}} &= \{f_f^0, f_d^1,f_f^1, f_d^2, f_f^2, ..., f_d^{m-1}, f_f^{m-1}\} + \textbf{P} + \textbf{T}
\end{split}
\end{equation}
$\mathbf{F_{te}}$ is the final temporal token output from temporal difference block, which is added with positional ($\textbf{P}$) and type ($\textbf{T}$) information. So, the frame tokens inserted with difference-enhanced tokens are input into temporal transformer, further promoting the sensitivity to capture motion-related information. Since $\mathbf{F_d}$ only describe the motion between frames, we only adopt output of frame tokens $\mathbf{F_v}= \{f_v^0, f_v^1, f_v^2, ..., f_v^{m-1}\}$ as video embedding, which consist of both spatial and temporal information. Then global average pooling is adopted to encode final video representation $f_v^g$.

\begin{figure}
        \begin{center}
            \includegraphics[width=1\linewidth]{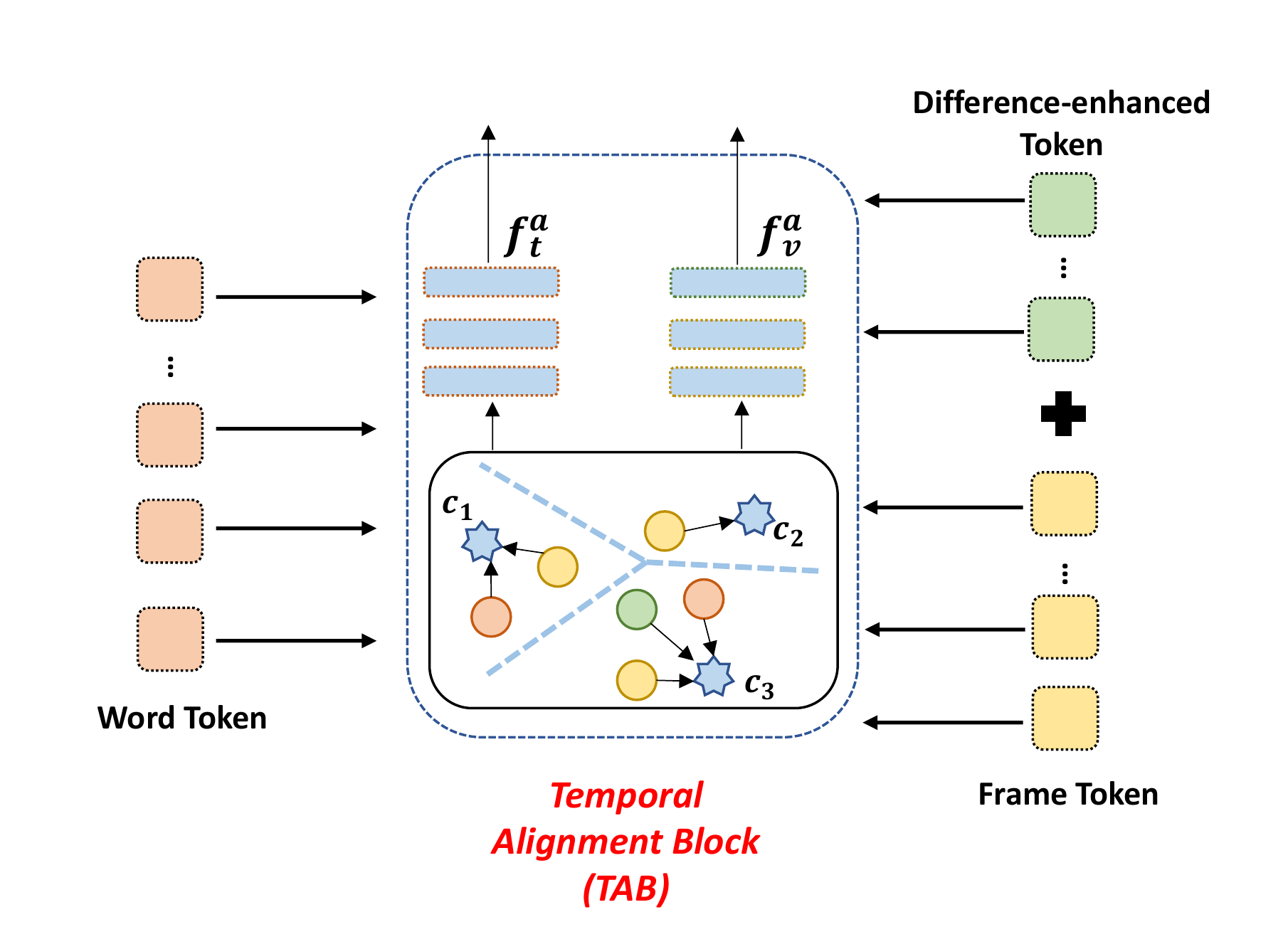}
        \end{center}
        \vspace{-0.2cm}
        \caption{Temporal Alignment Block. Word tokens and frame tokens with temporal-enhanced correlation are aligned into the shared centers $\textbf{C}$  for measuring the aggregated distribution.}
        \label{TAB}  
\end{figure}

\subsection{Temporal Alignment Block}
In common text-video retrieval, the modal representation is firstly calculated in individual domain, and then measure the similarity in joint space. By adopting temporal difference block with temporal transformer, the video embedding can be finely encoded. For text representation, 
we directly adopt CLIP's text encoder to generate the text representation, which is based on 12-layer transformer \cite{vaswani2017attention} modified by Radford \etal \cite{radford2019language}. Following CLIP \cite{radford2021learning}, lower-cased byte pair encoding (BPE) with a 49152 vocab size \cite{sennrich2015neural} is employed to tokenize input caption $\Phi$. The tokenized captions are bracketed with [CLS] and [SEP] token to indicate the start and end. Then text embedding is computed by the text encoder and representation can be seen as $\mathbf{F_{t}} = \{f_{t}^{cls}, f_{t}^0, f_{t}^1, ..., f_{t}^{n - 1}\}$, where $n$ is the sequence length. So, the output of [CLS] token which named as $f_{t}^{cls}$ is utilized as overall representation $f_t^g$ to minimize the distance with $f_v^g$ for global matching. However, since the existence of abundant contextual information in $\mathbf{F_{t}}$, which fully indicates the entire semantics, all word tokens can be adopted as auxiliary supervision to align key frames with apparent motion change.


Inspired by Netvlad \cite{arandjelovic2016netvlad}, we propose temporal alignment block to aggregate token embeddings of different modalities with shared centers and re-emphasise context-related representation. Specifically, $\mathfrak{K}$ shared centers $\{c_1,c_2, ..., c_k\}$ are learned to align frame and word embedding jointly. Following \cite{arandjelovic2016netvlad}, we calculate the confidence between modal features and shared centers by using dot-product, which is employed to assign weight for each cluster to measure the distribution. The formula is seen as follows:
\vspace{-0.2cm}
\begin{equation}
\begin{split}
\label{confidence}
\mathfrak{}{w}_{ij}=\frac{exp(\rho_ic_j^{\mathsf{T}} )}{\sum_{k=1}^K exp(\rho_ic_k^{\mathsf{T}} )}, \vspace{-0.2cm}
\end{split}
\end{equation}
where $\rho_i$ indicates $i$-th modal feature, $c_j$ is $j$-th shared center and $w_{ij}$ represents the normalized similarity. Then the aggregated embedding aligned with center $c_j$ can be obtained as:
\vspace{-0.2cm}
\begin{equation}
\begin{split}
\label{reweight}
\upsilon_j=\frac{\sum_{i=1}^\eta w_{ij}(\rho_i - \tilde{c}_j)}{\Vert \sum_{i=1}^\eta w_{ij}(\rho_i - \tilde{c}_j) \Vert_2}.
\end{split}
\end{equation}
$\eta$ is the max length of modal feature, and $\upsilon_j$ is $j$-th aligned center embedding. $\tilde{c}_j$ is the trainable weight that has the same size as $c_j$ to increase adaption \cite{arandjelovic2016netvlad}. Then center representation can be obtained as $\upsilon = \{\upsilon_1, \upsilon_2, ...,\upsilon_K\}$. Since the video and text are aggregated with shared centers of the same content, the overall semantic context in every modality token can be fully aligned into joint space before calculating the similarity. To further emphasize the weight of motion-related frame tokens toward action-described centers, we re-sample frame embedding sparsely with double frame rate from $\mathbf{F_f}$ as: $\mathbf{F_{fl}} = \{f_f^0, f_f^2, ...,f_f^{m-1}\}$. Although, $\mathbf{F_{fl}} $ sampled in large frame rate loses semantic coherence, it highlights changes in motion, which is beneficial as complementary information to re-adjust the weight distribution of motion-related center. To excite the temporal relationship, the shared temporal difference block is adopted to encode temporal tokens. We adopt a 1-layer transformer to correlate each temporal token and use the output of $\mathbf{F_{fl}} $ as $\mathbf{F_{dl}}$. Then $\mathbf{F_{dl}}$ is concatenated with $\mathbf{F_f}$ as $\mathbf{F_{ml}} = [\mathbf{F_f},\mathbf{F_{dl}}]$. Finally, aligned video and text representations can be seen as follows: 
\begin{gather}
\upsilon_j^{v}=\frac{1}{\sigma_v}\sum_{i=0}^{1.5(m-1)} \frac{exp(f_{ml}^ic_j^{\mathsf{T}} )}{\sum_{k=1}^K exp(f_{ml}^ic_k^{\mathsf{T}} )}(f_{ml}^i - \tilde{c}_j) \\
\upsilon_j^{t}=\frac{1}{\sigma_t}\sum_{i=0}^n \frac{exp(f_{t}^ic_j^{\mathsf{T}} )}{\sum_{k=1}^K exp(f_{t}^ic_k^{\mathsf{T}} )}(f_{t}^i - \tilde{c}_j),
\end{gather}
where $\sigma_v$ and $\sigma_t$ indicates the $l_2$ normalization. By adding the difference-enhanced frame tokens in large frame rate, the weight of action-described centers can be readjust for better alignment. Then global average pooling is also adopted to obtain final aligned representation $f^{a}_v$ and $f^{a}_t$ for video and text.

\subsection{Loss function}
To train CLIP2Video, we adopt symmetric cross entropy loss. Each training batch $\Omega$ consists of $B$ video-text pairs, which is discriminated in training:
\begin{gather}
\vspace{-0.2cm}
\small
\label{loss}
\mathcal{L}_{t2v}=- \frac{1}{B} \sum_{i \in \Omega} log\frac{exp(<f_t^i, f_v^i>)}{\sum_{j \in \Omega} exp(<f_t^i, f_v^j>)}, \\
\mathcal{L}_{v2t}=- \frac{1}{B} \sum_{i \in \Omega} log\frac{exp(<f_v^i, f_t^i>)}{\sum_{j \in \Omega} exp(<f_v^i, f_t^j>)}, \\
\mathcal{L}_{o} = \frac{1}{2}(\mathcal{L}_{t2v} + \mathcal{L}_{v2t})
\end{gather}
where $<f, f>$ indicates cosine similarity, and $\mathcal{L}_{o}$ is symmetric loss.
And, we adopt $f^g$ and $f^a$ to calculate $\mathcal{L}_{o}$ respectively.  So, the overall loss function can be seen as $\mathcal{L}_{o}=\frac{1}{2}(\mathcal{L}_{o}^{g}+\mathcal{L}_{o}^{a})$. The similarity during inference is formulated as: $<f_t, f_v> = \frac{1}{2}(<f_t^{g}, f_v^{g}> + <f_t^{a}, f_v^{a}>)$.

\section{Experiments}
\subsection{Experimental Settings}
\textbf{Datasets.} We conduct experiment on three benchmarks for  video-to-text retrieval and text-to-video retrieval tasks including MSR-VTT \cite{xu2016msr}, MSVD \cite{chen2011collecting} and VATEX \cite{wang2019vatex}. 
\begin{itemize}
    \item MSR-VTT \cite{xu2016msr} contains 10,000 videos, where each video contains 20 captions. We report result on 1k-A\cite{gabeur2020multi,miech2019howto100m,liu2019use} and full protocol \cite{dzabraev2021mdmmt,luo2020univilm} in our paper. Full protocol \cite{dzabraev2021mdmmt,luo2020univilm} is the standard split which includes 6513 videos for train, 497 videos for val and 2990 videos for test. In this protocol, each video contains multiple independent captions, which are all used in text-video retrieval. Besides, 1k-A protocol adopts 9,000 videos with all corresponding captions for training and utilizes 1,000 video-text pairs as test. When reporting the results of video-to-text retrieval, we adopt the maximum similarity among all corresponding captions for a given video query. 
    \item MSVD \cite{chen2011collecting} dataset includes 1,970 videos with approximately 80,000 captions, where train, validation and test are splited into 1200, 100 and 670 videos. In this paper, we report the results of test split with multiple captions per video.
    \item Besides, VATEX \cite{wang2019vatex} includes 34,991 videos with multilingual annotations. The training split contains 25,991 videos. Since it is inaccessible to obtain test annotation, we report the results following HGR's \cite{chen2020fine} validation protocol, which includes 1500 videos for validation and 1500 videos for test. For fair comparison, we adopt the English annotations.    
\end{itemize}

\textbf{Evaluation Metric.} We follow the standard retrieval task \cite{miech2018learning,mithun2018learning,zhang2018cross} and report Recall at rank K (R@K), median rank (MdR) and mean rank (MnR) as metric, where the higher R@K, and lower median rank and mean rank indicates better performance.

\textbf{Implementation Details.} 
We initialize the basic text transformer and spatial transformer (ViT) with CLIP (ViT-B/32). 
To initialize the other proposed transformers such as temporal transformer, we reuse parameters of similar dimensions in CLIP.  The dimension of all representations $f^g$ and $f^a$ in video and text is 512. The model is finetuned with Adam optimizer. The caption token length is 32 and frame length is 12 in our settings. Meanwhile the length of layer in video transformer are 12 and 4 for $L_s$ and $L_t$. Besides, the shared centers $c$ and $\tilde{c}$ are the trainable weight of $K \times 512 $ dimension. And, we adopt $K$=5 for training in MSVD and MSR-VTT and set $K$=7 in VATEX for temporal alignment block. More thorough implementation details can be found in supplemental material.

\begin{table*}

	\begin{center}
		\resizebox{0.85\textwidth}{!}{
			\begin{tabular}{c|ccccc|ccccc} 
        	\hline
            \multicolumn{1}{c}{} & \multicolumn{5}{c}{Text $\Longrightarrow$ Video} &\multicolumn{5}{c}{Video $\Longrightarrow$ Text} \\
            \hline
        	Method & R@1 & R@5 & R@10 & MdR & MnR & R@1 & R@5 & R@10 & MdR & MnR \\
            \hline
        Mean Pooling   \cite{luo2021clip4clip}              & 43.1 & 70.4 & 80.8 & 2.0 & 16.2 & 43.1 & 70.5 & 81.2 & 2.0 & 12.4 \\
         Temporal Transformer \cite{luo2021clip4clip}   & 44.5 & 71.4 & 81.6 & 2.0 & 15.3 & 42.7 & 70.9 & 80.6 & 2.0 & 11.6 \\
         TDB-Sub     & 44.2 & 70.8 & 80.7 & 2.0 & 15.7 & 41.9  & 70.1 & 80.6 & 2.0 & 11.7  \\
        TDB-MLP             & 44.7 & 70.9    & 80.9 & 2.0    & 14.6 & 42.7  & 70.8     & 81.4 &2.0 & 11.3  \\
        TDB-All             & 44.3 & 71.3 & 82.1 & 2.0 & 15.6 & 41.8 & 69.9 & 80.8 & 2.0 & 11.9 \\
         \hline
        \textbf{TDB}      & \textbf{45.1} & \textbf{72.4} & \textbf{80.7} & \textbf{2.0} & \textbf{14.6} & \textbf{41.9} & \textbf{71.1} & \textbf{80.3} & \textbf{2.0} & \textbf{10.8} \\
        \hline
         TDB+TAB-Base & 45.1 & 72.7 & 80.9 & 2.0 & 15.1 & 42.4 & 70.8 & 81.9 & 2.0 & 10.0 \\
        TDB+TAB-Temporal   & 44.1 & 72.7 & 82.2 & 2.0 & 14.2 & 43.1 & 71.4 & 82.1 & 2.0 & 10.2 \\
         TDB+TAB-Transformer   & 44.6 & 72.5 & 82.0 & 2.0 & 13.7 & 42.3 & 71.9 & 82.9 & 2.0 & 9.8 \\
         \hline
         \textbf{TDB+TAB-TDB (ours)}    & \textbf{45.6} & \textbf{72.6} & \textbf{81.7} & \textbf{2.0} & \textbf{14.6} & \textbf{43.3} & \textbf{72.3} & \textbf{82.1} & \textbf{2.0} & \textbf{10.2} \\
        	\hline
	        \end{tabular}}
	\end{center}
	\vspace{-0.2cm}
	\caption{Comparative results with different settings of temporal difference block (TDB) and temporal alignment block (TAB). The results of 1k-A protocol in MSR-VTT are reported.}
	\label{MSRVTTABlation}
\end{table*}

  \begin{table*}
	\begin{center}
		\resizebox{0.85\textwidth}{!}{
			\begin{tabular}{c|ccccc|ccccc} 
        	\hline
            \multicolumn{1}{c}{} & \multicolumn{5}{c}{Text $\Longrightarrow$ Video} &\multicolumn{5}{c}{Video $\Longrightarrow$ Text} \\
            \hline
        	Method & R@1 & R@5 & R@10 & MdR & MnR & R@1 & R@5 & R@10 & MdR & MnR \\
            \hline
            TAB-center=4 & 44.6 & 72.9 & 82.2 & 2.0 & 13.7 & 43.7 & 72.5 & 82.8 & 2.0 & 10.3 \\
            TAB-center=6 & 44.6 & 71.8 & 81.3 & 2.0 & 14.4 & 44.7 & 72.8 & 82.5 & 2.0 & 10.2 \\
            TAB-center=7 & 44.1 & 72.8 & 81.3 & 2.0 & 15.1 & 42.3 & 71.2 & 82.0 & 2.0 & 10.4 \\
        	\hline
        	\hline
           TAB-weight=0.4 & 44.1 & 72.9 & 82.5 & 2.0 & 14.4 & 43.7 & 72.8 & 81.2 & 2.0 & 10.1 \\
           TAB-weight=0.6 & 44.9 & 72.1 & 82.7 & 2.0 & 14.4 & 43.3 & 72.6 & 81.9 & 2.0 & 10.3 \\
           TAB-weight=0.7 & 43.7 & 72.2 & 81.7 & 2.0 & 14.1 & 41.8 & 70.7 & 82.4 & 2.0 & 10.4 \\
        	\hline
	        \end{tabular}}
	\end{center}
	\caption{The results of different weights of temporal alignment block (TAB) on MSR-VTT. }
	\label{weight}
\end{table*}

\subsection{Ablation Studies}
\subsubsection{Effects of Temporal Difference Block.} Compared with mean pooling, the usage of temporal transformer achieves better performance by interacting and aggregating frames. To further enhance the temporal correlation, we adopt the difference of adjacent frame embedding as description of action to insert into the frame tokens. Specifically, difference-level attention (1-layer transformer) is adopted to encode the subtraction as difference. 
For fair comparison, we report the results of different settings in Tab.\ref{MSRVTTABlation}, exploiting the best type of video representation. It can be seen that inserting subtraction directly (\textit{TDB-Sub}) or just employing MLP (\textit{TDB-MLP}) to encode correlations, the whole frame representation with positional indices will be damaged.
The implicit difference with weak transformation increase the difficulty of temporal transformer to capture the motion-related information. Instead, the usage of  difference-level attention (\textit{TDB}) provides the explicit interaction before inserting, which significantly improves the performance. Besides, we also exploit to whether adopt global average pooling on all the output of tokens. Since the lack of complete spatial information, adopting all the output of tokens (\textit{TDB-All}) including difference-enhanced tokens, inevitably downgrades the retrieval performance evidently.

\begin{table*}[t]
	\begin{center}
		\resizebox{0.85\textwidth}{!}{
		\begin{tabular}{c|c|ccccc|ccccc} 
	    \hline
        \multicolumn{1}{c}{} & \multicolumn{1}{c}{} & \multicolumn{5}{c}{Text $\Longrightarrow$ Video} &\multicolumn{5}{c}{Video $\Longrightarrow$ Text} \\
        \hline
	    Method & Test-set & R@1 & R@5 & R@10 & MdR & MnR & R@1 & R@5 & R@10 & MdR & MnR \\
        \hline
	    JSFusion \cite{yu2018joint}             & 1k-A & 10.2 & 31.2 & 43.2 & 13.0 & -    & -    &    - & -    &   - & - \\
	    HT-pretrained \cite{miech2019howto100m} & 1k-A & 14.9 & 40.2 & 52.8 & 9.0  & -    & -    &    - & -    &   - & - \\
	    CE \cite{liu2019use}                    & 1k-A & 20.9 & 48.8 & 62.4 & 6.0  & 28.2 & 20.6 & 50.3 & 64.0 & 5.3 & 25.1 \\
	    MMT-Pretrained \cite{gabeur2020multi}   & 1k-A & 26.6 & 57.1 & 69.6	& 4.0  & 24.0 & 27.0 & 57.5 & 69.7 & 3.7 & 21.3 \\
	    SUPPORT-SET \cite{patrick2020support}   & 1k-A & 27.4 & 56.3 & 67.7 & 3.0  & -    & 26.6 & 55.1 & 67.5 & 3.0 & - \\
	    FROZEN \cite{bain2021frozen}            & 1k-A & 31.0 & 59.5 & 70.5	& 3.0  & -    & -    &    - & -    &   - & - \\
	    CLIP \cite{portillo2021straightforward} & 1k-A & 31.2 & 53.7 & 64.2 & 4.0  & -    & 27.2 & 51.7 & 62.6 & 5.0 & - \\
	    HIT-pretrained \cite{liu2021hit}        & 1k-A & 30.7 & 60.9 & 73.2 & 2.6  & -    & 32.1 & 62.7 & 74.1 & 3.0 & - \\
	    MDMMT \cite{dzabraev2021mdmmt}          & 1k-A & 38.9 & 69.0 & 79.7	& 2.0  & 16.5 & -    &    - & -    &   - & - \\
        CLIP4Clip-meanP \cite{luo2021clip4clip}& 1k-A & 43.1 & 70.4 & 80.8	& 2.0  & 16.2 & 43.1 & 70.5 & 81.2 & 2.0 & 12.4 \\	
        CLIP4Clip-seqTransf \cite{luo2021clip4clip}& 1k-A & 44.5 & 71.4 & 81.6	& 2.0  & 15.3 & 42.7 & 70.9 & 80.6 & 2.0 & 11.6 \\	
	    \hline
	    \textbf{ours} & 1k-A & \textbf{45.6} & \textbf{72.6} & \textbf{81.7} & \textbf{2.0} & \textbf{14.6} & \textbf{43.5} & \textbf{72.3} & \textbf{82.1} & \textbf{2.0} & \textbf{10.2} \\
	    \hline
	    Dual Enc. \cite{dong2019dual}           & Full & 7.7  & 22.0 & 31.8 & 32.0 & -    & 13.0 & 30.8 & 43.3 & 15.0& - \\
	    E2E \cite{miech2020end}                 & Full & 9.9  & 24.0 & 32.4 & 29.5 & -    & -    &    - & -    &   - & - \\
	    CE \cite{liu2019use}                    & Full & 10.0 & 29.0 & 41.2 & 16.0 & 86.8 & 15.6 & 40.9 & 55.2 & 8.3 & 38.1 \\
	    HT-pretrained \cite{miech2019howto100m} & Full & 14.9 & 40.2 & 52.8 & 9.0  & -    & -    &    - & -    &   - & - \\
	    CLIP \cite{portillo2021straightforward} & Full & 21.4 & 41.1 & 50.4 & 10.0 & -    & 40.3 & 69.7 & 79.2 & 2.0 & - \\
	    UNiVL \cite{luo2020univilm}             & Full & 21.2 & 49.6 & 63.1 & 6.0  & -    & -    &    - & -    &   - & - \\	
	    MDMMT \cite{dzabraev2021mdmmt}          & Full & 23.1 & 49.8 & 61.8	& 6.0  & 52.8 & -    &    - & -    &   - & - \\
	    \hline
	    \textbf{ours}  & 1k-A & \textbf{29.8} & \textbf{55.5} & \textbf{66.2} & \textbf{4.0}  & \textbf{45.5} & \textbf{54.6} & \textbf{82.1} & \textbf{90.8} & \textbf{1.0} & \textbf{5.3} \\
    	\hline
		\end{tabular}}
	\end{center}
	\vspace{-0.2 cm}
	\caption{Retrieval result on MSR-VTT. \textbf{1k-A} indicates test set of 1000 pairs used by \cite{yu2018joint}, while \textbf{full} represents the standard test set. CLIP4Clip-meanP and CLIP4Clip-seqTransf indicate the version with mean pooling and temporal transformer for frame aggregation.}
	\label{MSRVTTTable}
\end{table*}

\begin{table*}
	\begin{center}
		\resizebox{0.85\textwidth}{!}{
		\begin{tabular}{c|ccccc|ccccc} 
	    \hline
        \multicolumn{1}{c}{} & \multicolumn{5}{c}{Text $\Longrightarrow$ Video} &\multicolumn{5}{c}{Video $\Longrightarrow$ Text} \\
        \hline
	    Method & R@1 & R@5 & R@10 & MdR & MnR & R@1 & R@5 & R@10 & MdR & MnR \\
        \hline
        VSE \cite{kiros2014unifying}                  & 12.3 & 30.1 & 42.3    & 14.0 & -     & 34.7 & 59.9 & 70.0 & 3.0 & - \\
        CE \cite{liu2019use}                          & 19.8 & 49.0 & 63.8    & 6.0  & 23.1  & -    &    - & -    &   - & - \\
        SSML \cite{amrani2020noise}                   & 20.3 & 49.0 & 63.3	  & 6.0  & -     & -    &    - & -    &   - & - \\
        SUPPORT-SET \cite{patrick2020support}         & 28.4 & 60.0 & 72.9    & 4.0  & -     & -    &    - & -    &   - & - \\
        FROZEN \cite{bain2021frozen}                  & 33.7 & 64.7 & 76.3	  & 3.0  & -     & -    &    - & -    &   - & - \\
        CLIP \cite{portillo2021straightforward}       & 37.0 & 64.1 & 73.8    & 3.0  & -     & 59.9 & 85.2 & 90.7 & 1.0 & - \\
        CLIP4Clip-seqTransf \cite{luo2021clip4clip}   & 45.2 & 75.5 & 84.3	  & 2.0  & 10.0  & \textbf{62.0} & \textbf{87.3} & \textbf{92.6} & 1.0 & 4.3 \\
        CLIP4Clip-meanP \cite{luo2021clip4clip}       & 46.t2 & 76.1 & 84.6	  & 2.0  & 10.0  & 56.6 & 79.7 & 84.3 & 1.0 & 7.6 \\
	    \hline
	    \textbf{ours} & \textbf{47.0} & \textbf{76.8} & \textbf{85.9} & \textbf{2.0}  & \textbf{9.6} & 58.7	& 85.6 & 91.6 & \textbf{1.0} & \textbf{4.3} \\
	    \hline
		\end{tabular}}
	\end{center}
	\vspace{-0.2cm}
	\caption{Retrieval results on MSVD.}
	\label{MSVDTable}
\end{table*}

\begin{table*}
	\begin{center}
		\resizebox{0.85\textwidth}{!}{
		\begin{tabular}{c|ccccc|ccccc} 
        \hline
        \multicolumn{1}{c}{} & \multicolumn{5}{c}{Text $\Longrightarrow$ Video} & \multicolumn{5}{c}{Video $\Longrightarrow$ Text} \\
        \hline
        Method & R@1 & @5& @10& MdR & MnR & @1& @5& @10& MdR & MnR \\
        \hline
        Random Baseline                               & 0.2  & 0.7  & 1.05    & 2000.5 & -   & 0.02 & 0.1 & 1.02 & 2100.5 & -\\
        VSE \cite{kiros2014unifying}                  & 28.0 & 64.3 & 76.9    & 3.0    & -   & -    &    - & -    &   -   & - \\
        SE++                                    & 33.7 & 70.1 & 81.0    & 2.0    & -   & -    &    - & -    &   -   & - \\
        Dual Enc. \cite{dong2019dual}                 & 31.1 & 67.5 & 78.9    & 3.0    & -   & -    &    - & -    &   -   & - \\
        HGR   \cite{chen2020fine}                     & 35.1 & 73.5 & 83.5    & 2.0    & -   & -    &    - & -    &   -   & - \\
        CLIP \cite{portillo2021straightforward}       & 39.7 & 72.3 & 82.2    & 2.0    & 12.8& 52.7 & 88.8 & 94.9 & 1.0   & 3.8 \\
        SUPPORT-SET \cite{patrick2020support}         & 44.9 & 82.1 & 89.7    & 1.0    & -   & 58.4 & 84.4 & 91.0 & 1.0   & - \\
        CLIP4Clip-seqTransf \cite{luo2021clip4clip}   & 55.9 & 89.2 & 95.0    & 1.0    & 3.9 & 73.2 & 97.1 & 99.1 & 1.0   & 1.7 \\
        \hline
        \textbf{ours} & \textbf{57.3} & \textbf{90.0} & \textbf{95.5} & \textbf{1.0} & \textbf{3.6} & \textbf{76}& \textbf{97.7} &\textbf{99.9}	& \textbf{1.0} & \textbf{1.5} \\
        \hline
	    \end{tabular}}
	\end{center}
	\vspace{-0.2cm}
	\caption{Retrieval results on VATEX. }
	\label{VATEXTable}
\end{table*}

\subsubsection{Effects of Temporal Alignment Block.} We adopt frame embedding concatenated with difference-enhanced embedding in double frame rate to aggregate text embedding by aligning the shared centers. In this section, we compare different types of difference-enhanced embedding for alignment and also report the results in Tab.\ref{MSRVTTABlation}. By introducing basic alignment (\textit{TDB+TAB-base}), the performance of video-to-text retrieval has achieved the evident improvement. However, since each token embedding of text contain abundant contextual information, it is hard to aggregate independently modeled frame representation, due to the semantic gap. One way to solve that is to utilize the shared temporal output (\textit{TDB+TAB-Temporal}) as difference-enhanced embedding, but the weight of motion-related frames can not be strengthened, since every token contains the whole interaction. Instead, we adopt frame embedding for basic alignment, and add extra frame embedding with 1-layer transformer (\textit{TDB+TAB-Transformer}).
The representations with large frame rate encode the apparent motion change in less frames, re-distributing the weight of motion-related center to align with the context of text. Meanwhile, adopting TDB (\textit{TDB+TAB-TDB}) to further encode the temporal relationship in large frame rate, we have achieve the best performance.

We also give more ablation studies to exploit the settings of hyper-parameters. In Tab.\ref{weight}, we compare the results of different center numbers $K$ on MSR-VTT for alignment, based on the results of \textit{TDB+TAB-base}. The results in Tab.\ref{weight} shows that performance degrades with the evidently increase of $K$.  Since the limited number of videos in MSR-VTT, it is hard for convergence when aligning with large number of centers $K$. So, we choose $K=5$ as the best settings in MSVD and MSR-VTT. Besides, due to the more number of videos in VATEX, we adopt $K=7$ to provide more centers to finely discriminate the key frames. When calculating the loss during training, we adopt $\mathcal{L}_{o}=w(\mathcal{L}_{o}^{g})+(1-w)\mathcal{L}_{o}^{a}$, where $w$ is 0.5 in our paper. During inference, the similarity between video and text is formulated as: $<f_t, f_v> = w(<f_t^{g}, f_v^{g}>) + (1-w)(<f_t^{a}, f_v^{a}>)$, where $w$ is $\frac{1}{2}$. To exploit the weight of two similarities, we give the results of different $w$ and report in Tab.\ref{weight}. With the increase of weight, the confidence of aligned similarity has been weakened, which damages the whole representation performance. Since the alignment 
is not sufficiently trained with low weight $w$, the retrieval performance of only adopting TDB is better than adopting TDB and TAB simultaneously. However, with the large weight of alignment, loss also can not be converged well, due to the limited initialization. Since both $f^g$ and $f^l$ are equally important, we adopt $w$=0.5 in our paper to achieve the best performance.

\begin{figure*}
    \begin{center}
        \includegraphics[width=0.85\linewidth]{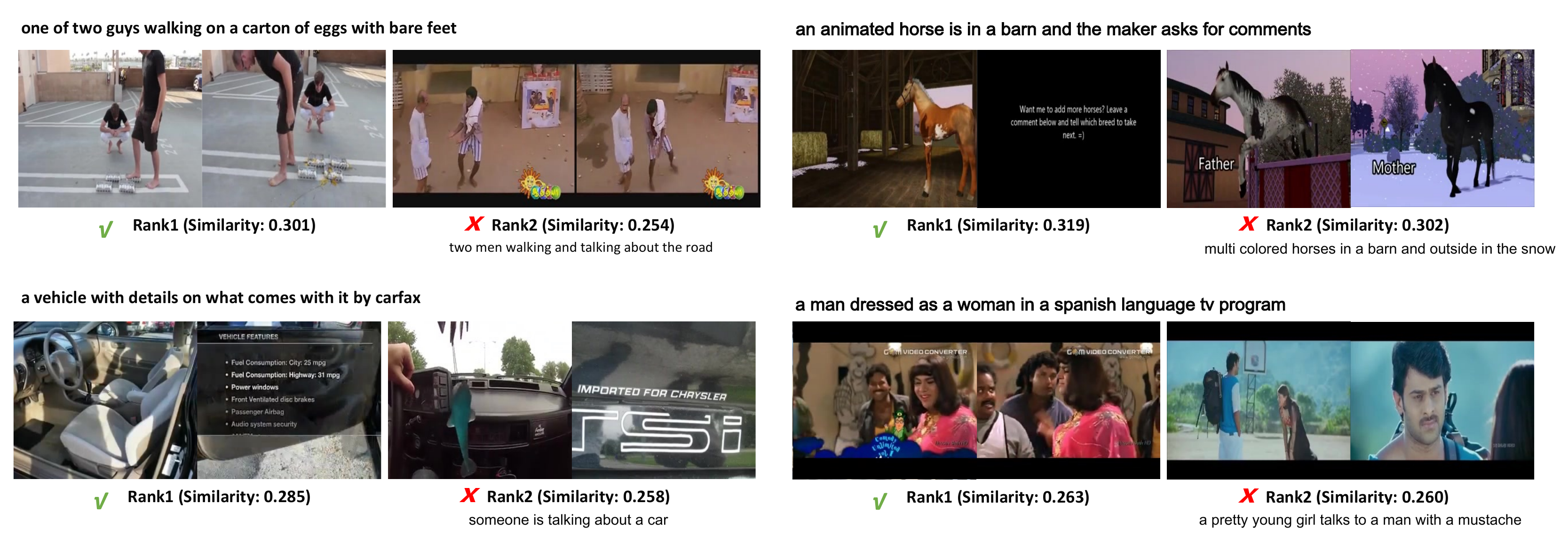}
    \end{center}
    \vspace{-0.7cm}
    \begin{center}
        \includegraphics[width=0.85\linewidth]{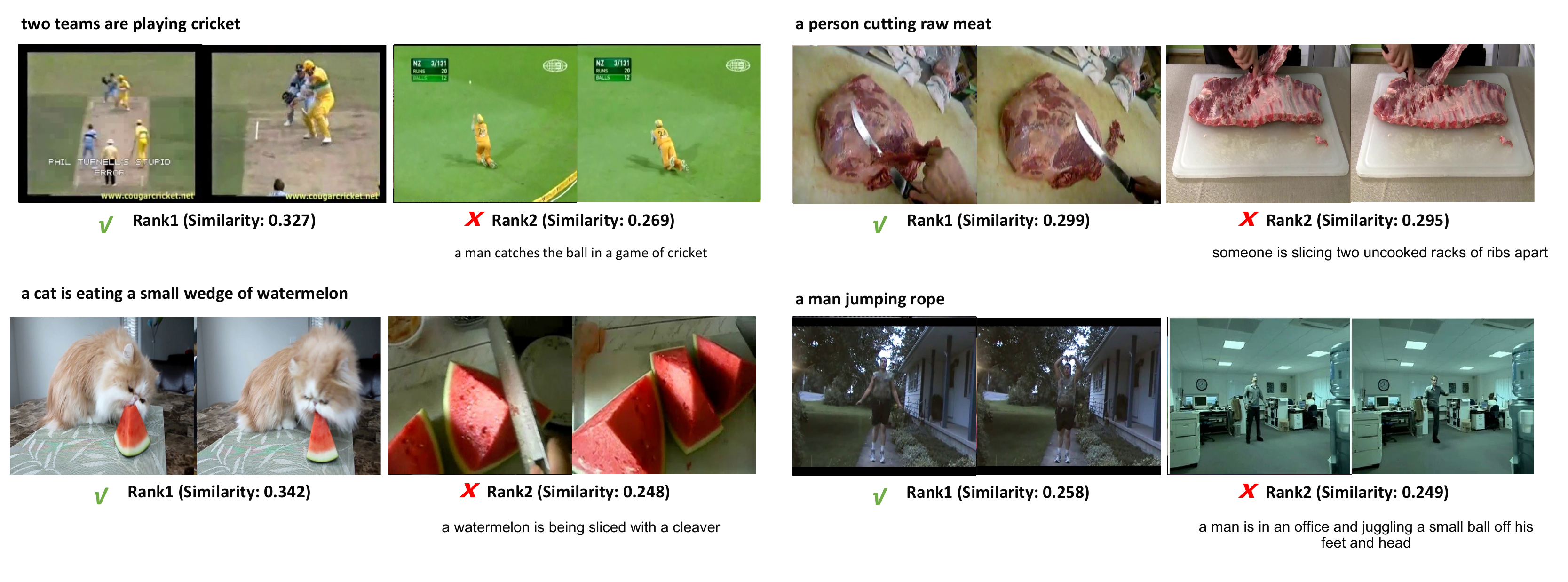}
    \end{center}
    \vspace{-0.7cm}
    \begin{center}
        \includegraphics[width=0.85\linewidth]{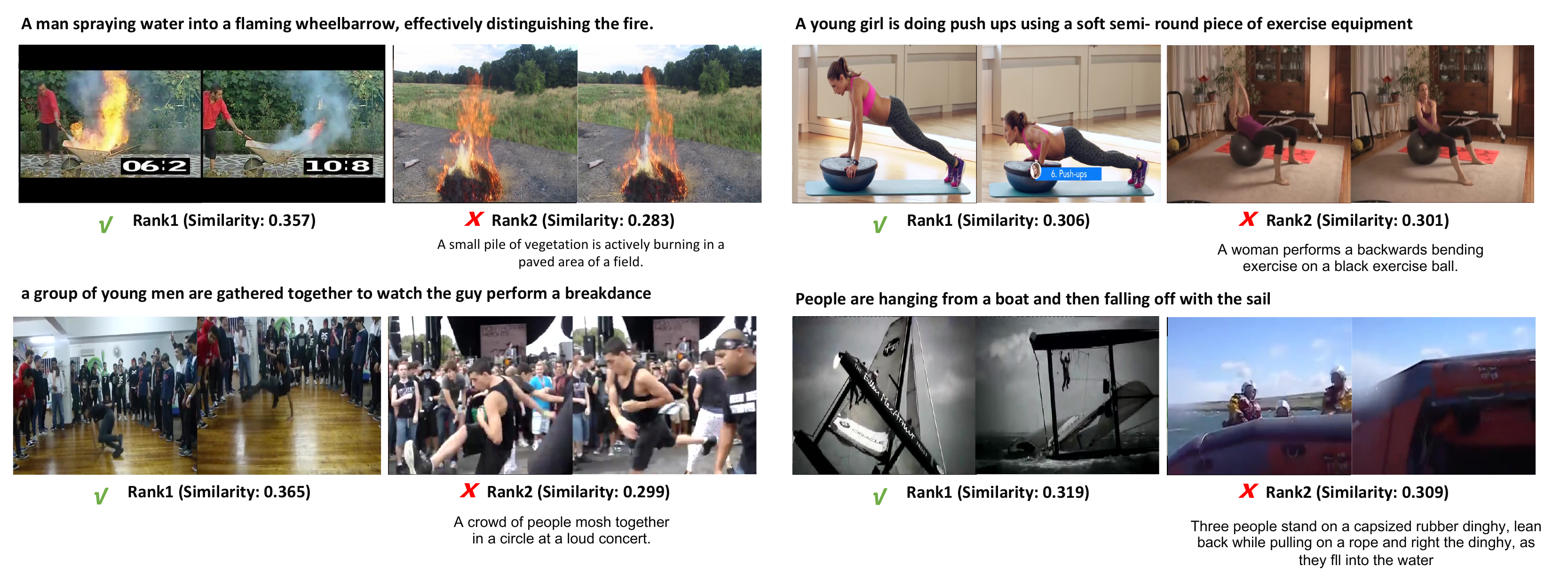}
    \end{center}
    \vspace{-0.6cm}
    \caption{ 
    The text-to-video retrieval results on MSR-VTT, MSVD and VATEX. The upper-left is the query caption for each group. And each two frames are sampled from the target video. Besides, the correct caption of videos in rank 2 is also given in the bottom.
    }
    \label{vis}        
\end{figure*}

\subsection{Comparison with Other Methods}
We compare our proposed method against the state-of-the-art. All the results of video-to-text and text-to-video retrieval are reported in Tab.\ref{MSRVTTTable},\ref{MSVDTable},\ref{VATEXTable}. We achieve the SOTA results on all three datasets compared with baselines, where the visible growth of performance can be found by employing CLIP \cite{portillo2021straightforward} as pre-trained model. Benefit from the large-scale image-text pre-training,  zero-shot retrieval of CLIP with mean pooling has surpassed most of the fine-tuning work. Although, the usage of CLIP doesn't fully exploit the temporal relationship of video, the powerful spatial representations in short video alleviate the lack of information. Meanwhile, adopting CLIP for fine-tuning and modeling interaction between frames with temporal transformer, the performance can be further increased, which is shown by CLIP4clip \cite{luo2021clip4clip} on Tab.\ref{MSRVTTTable}. 
However, when give a small dataset such as MSVD, it is hard to learn temporal representation with temporal transformer by only relying positional embedding to indicate temporal indices. Instead, we insert temporal-enhanced tokens between frame tokens to guide temporal transformer to caption dynamic motion patterns.
And, the whole contextual words are utilized to align with key frame with abundant semantics by the shared centers. So, the optimization can be supervised to focus on fusing temporal representation, achieving better performance even in small dataset. 

\subsection{Qualitative Results}
We show two kinds of videos retrieved by our proposed method.  As depicted in Fig.\ref{vis}, The left two queries demonstrates easy results with large margin of similarity. Since, the scene of video conveys the evident difference, our CLIP2video can will retrieve them by introducing more temporal information to describe the action precisely. Besides, we also give some hard samples, which are shown in the right of Fig.\ref{vis}, where it is hard to discriminate them due to the similar frame. For example, when give query: "an animated horse is in a barn and the maker asks for comments", the searched videos with high confidence both contain the animated horse in the barn. However, the half of capitation emphasizes that " the maker asks for comments", which is aligned to the specific centers to aggregate target frames. So, the weight of frames including subtitles can be enhanced with temporal alignment block and help to retrieve the video that best meets the description.

\section{Conclusion}
We redefine the video-text retrieval from a macroscopic view of point, dividing it into a image-text multi-modal learning and temporal relationships between video frames and video-text. Aiming to consider both sides, we propose CLIP2Video network to transfer the image-language pre-training model to video-text retrieval, which based on a image-language pretraining model and two Temporal Blocks to capture motions at fine temporal frames and re-align the tokens between video and languages respectively. Our experimental results show that the proposed approach can significantly improve the performance on several text-video retrieval benchmarks, including new records on MSR-VTT, MSVD, VATEX.

{\small
\bibliographystyle{ieee_fullname}
\bibliography{egbib}
}

\end{document}